%% file: Artigo.tex
\documentclass[12pt]{article}

\usepackage{sbc-template}
\usepackage{babel}
\usepackage[T1]{fontenc}
\usepackage[utf8]{inputenc}
\usepackage{multicol}
\usepackage{multirow}
\usepackage{graphicx,url}
\usepackage{amsmath,amssymb,amsfonts}
\usepackage{algorithmic}
\usepackage{textcomp}
\usepackage{enumerate}
\usepackage{soul}
\usepackage{verbatim}
\usepackage[table,xcdraw]{xcolor}
\usepackage[shortlabels]{enumitem}
\usepackage{scalefnt}
\usepackage{comment} 
\usepackage{lscape}
\usepackage{fancyhdr}
\usepackage{sbgames} 
\usepackage[super]{nth}

\emergencystretch=\maxdimen
\input{sbgames_info}

\title{SpellForger: Prompting Custom Spell Properties In-Game using BERT supervised-trained model}

\author{Emanuel C. Silva\inst{1*}, Emily S. M. Salum\inst{1*}, Gabriel M. Arantes\inst{1*}, \\ Matheus P. Pereira\inst{1*} \and Vinicius F. Oliveira\inst{1}\thanks{These authors contributed equally to this work.}  \and Alessandro L. Bicho\inst{1}}%

\address{
Centro de Ciências Computacionais -- Universidade Federal do Rio Grande (FURG)\inst{1}\\
Rio Grande -- RS -- Brasil\\}

\begin{document}

\maketitle
\begin{center}
    \texttt{\{emanuel\_silva, emilymarquessalum, gma26062004, matheusporto212, 12viniciusfoliveiraa, albicho\}@furg.br}%
\end{center}


\thispagestyle{plain}

\begin{abstract}

\textit{Introduction:} The application of Artificial Intelligence in games has evolved significantly, allowing for dynamic content generation. However, its use as a core gameplay co-creation tool remains underexplored. \textit{Objective:} This paper proposes \textbf{SpellForger}, a game where players create custom spells by writing natural language prompts, aiming to provide a unique experience of personalization and creativity. \textit{Methodology:} The system uses a supervised-trained BERT model to interpret player prompts. This model maps textual descriptions to one of many spell prefabs and balances their parameters (damage, cost, effects) to ensure competitive integrity. The game is developed in the Unity Game Engine, and the AI backend is in Python. \textit{Expected Results:} We expect to deliver a functional prototype that demonstrates the generation of spells in real time, applied to an engaging gameplay loop, where player creativity is central to the experience, validating the use of AI as a direct gameplay mechanic.
\end{abstract}


\section{Introduction}

The application of Artificial Intelligence (AI) in games has been used in the past for several applications, most often for Non-Player Characters (NPC) behavior, with simple methods such as Finite-State Machines \cite{jagdale2021finite}, and in recent years more complex systems using Machine Learning \cite{inproceedings}. This work presents a project called \textbf{SpellForger}, a game that advances in this direction by using Natural Language Processing (NLP) as a central playability mechanic.

The main concept of \textbf{SpellForger} is the freedom the players have to create their own spells by writing a description in a prompt. Then, a Large Language Model (LLM) interprets it and generates a unique spell for the player. This system aims to provide an unprecedented level of personalization and creativity, where the player's strategy begins at its own spell forging.

The remainder of the paper is organized as follows. Section II presents a literature review. The methodology and game proposal are presented in Section III, and the gameplay is detailed in Section IV. Finally, in Section V, some final remarks are made, and directions for future research are indicated.


\section{Literature Review}
 
Procedurally Generated Content (PGC) in games is not something new; titles like \textbf{No Man's Sky} \cite{HelloGames2016NoMansSky} and \textbf{Minecraft} \cite{Persson2010Minecraft} use it to create diverse and large worlds. \textbf{ChatGepetto} \cite{sbgamesChatGpt} uses GPT, a generative model, to create game narratives in real time. In  \cite{sbgamesMusicAI}, a transformer deep learning model is used to generate music in real time, with the intention of creating personalized experiences for different users.
In \cite{9720203}, a BERT model is used to generate \textbf{Minecraft} world snippets.  
Bidirectional Encoder Representations from Transformers (BERT) \cite{usherwood2019lowshotclassificationcomparisonclassical} is a model that takes text and encodes it into a representation for other models. 
We can then add a layer that takes the result of the model and turns it into desired values, by the process called fine-tuning, so it can adapt to any specific task.

Our project differentiates itself by applying those technologies not in narrative or landscape generation, but in the game's central mechanic interactions.
Games like \textbf{Galactic Arms Race} \cite{articleGalaticArms} have attempted to personalize the user experience in the central mechanic by understanding their preferences with machine learning, but this method is different from ours as it does not give the user full control of what's available to them. Rather than giving players options and trying to understand their preferences, our focus is on allowing players to access any option they wish through their expression, allowing them to explore and learn how our system reacts to their interaction. 
The use for interpreting prompts and turning them into parametrized abilities represents a novel approach to playability personalization.

\section{Methodology and game proposal}

This section explains our work in more detail, within the technological aspects of the proposed system.

\subsection{Architecture and technologies}

Our development used the Unity Game Engine,
 chosen due to its robustness and ample support for 3D and multiplayer games. The AI logic to the spell generation was developed in Python, using machine learning libraries (PyTorch \cite{pytorch} and Scikit-learn \cite{scikit-learn}). For the language model, we used a BERT-based architecture, where the model is trained like a supervised model, with a dataset of spell descriptions and their respective numerical and functional attributes, making it a bridge between human language and game logic. The BERT model was chosen due to its speed after training, being more lightweight than a generative model like GPT-2 would be.

\subsection{Central Mechanic: AI Spell Generation} \label{sec: spell generation}

The innovative part of our project resides in the magic generation, where the AI acts as an interpreter and balancer of the forged spells. 
Our methodology of using a BERT model meant we would require to define a spell within a fixed amount of resulting parameters. In order to define a system that could generate dynamic behavior, we had to choose relevant values that could allow for variety. Thus, the characteristics of a spell were chosen to be: its type, its status, and its effects. 

When a player starts a new match, they must insert a prompt. For this section, we will use this example of input a user might type: "A trap that holds the enemy to the ground." Unity will then communicate with Python code through running a script, which imports the language model that will analyze and associate the description with a spell type from the available types defined for the system. It will also adjust the status parameters that will further customize the spell and generate the status effects for that spell. The Unity system will then receive those results and map them to a specific spell prefab, which will reconfigure itself based on the parameters. Finally, the Trigger scripts will be attached to the game object of the spell based on the status effects (Section \ref{sec: spell effect}). Any other triggers will also be incorporated when they are fundamental mechanics of that spell type. The complete process, which runs only on the computer of the user, takes on average 200ms, but further studies must be done to arrive at precise measurements taking into account different hardware setups.

\subsubsection{Spell Type}
During the development, we used Unity's Prefab system\footnote{Unity Technologies. "Prefabs." \textit{Unity Manual}. Available: \url{https://docs.unity3d.com/6000.1/Documentation/Manual/Prefabs.html}} to define different spells. A valid spell prefab would have a script extending from \textit{SpellPrefabType}.
Each spell prefab is added to a list inside a \textit{SpellTypeController}, where it also must have a name assigned to it. It's important to note the name informs the developers what to expect in the visual of the spell and how it will interact, but it does not teach the model directly nor is it informed to the player.   
For the initial version, we implemented these types: Projectile, Fireball, Thunder, Trap, and Area Effect. Other types will eventually be added to bring more diversity to the gameplay. 
In the system, a spell whose type is set to '0' would be mapped as a Projectile, while a type '1' would be a Fireball and so on. The example (Section \ref{sec: spell generation}) would generate the value '3' for type.

\subsubsection{Spell Status}
We defined status as numerical values that alter the base form of a spell. They were: Power, Speed, Area, and Color.
The use of these statuses varies from type to type, where each type of spell is tasked to choose what a status will do. For example, Speed will change the velocity of a projectile spell, but a stationary spell could not use it for that purpose.
Each feature has a maximum value.
The immediately generated status is not used directly, as it would restrict the developer's agency in tweaking how the status scales. Instead, we limit the value of parameters using linear interpolation of pre-defined ranges. Each status has a sequence: $V = \{v_0, v_1, v_2, \ldots, v_{n-1}\}$.
We can define its real value as $V_{\text{real}} = v_i + (v_{i+1} - v_i) \cdot j$, where $i$ is the integer part of the value generated by our model, and $j$ is the fractional part of the value.

\subsubsection{Spell Effect} \label{sec: spell effect}
The effects of a spell are defined through a \textbf{Status Effects Matrix}, denoted as $M$. This matrix is structured as a $4 \times 4$ array of integer values, where each element $M_{i,j}$ quantifies the intensity or nature of a status effect applied under specific trigger conditions.

The rows of the matrix, indexed by $i \in \{0, 1, 2, 3\}$, correspond to distinct triggers: $T_0$ for a collision with an enemy, $T_1$ for a collision with any player, $T_2$ for a collision with an ally, and $T_3$ for all entities affected by the spell's general area or effect. The columns, indexed by $j \in \{0, 1, 2, 3\}$, represent the statuses that can be affected: $S_0$ for Health, $S_1$ for Speed, $S_2$ for Defense, and $S_3$ for Mana.

Each element $M_{i,j}$ can take a value from the range $[-1, 1]$. A value of $0$ indicates no effect. Non-zero values ($1$ or $-1$) denote the application of a positive or negative status modifier, respectively, upon the activation of trigger $T_i$ on status $S_j$. For instance, a value of $-1$ for $M_{0,0}$ implies a damage-over-time effect (health debuff) on an enemy upon collision.

$$M = \begin{pmatrix}
M_{0,0} & M_{0,1} & M_{0,2} & M_{0,3} \\
M_{1,0} & M_{1,1} & M_{1,2} & M_{1,3} \\
M_{2,0} & M_{2,1} & M_{2,2} & M_{2,3} \\
M_{3,0} & M_{3,1} & M_{3,2} & M_{3,3}
\end{pmatrix}$$
where $M_{i,j} \in [-1, 1]$.

\subsubsection{Spell cost}
In order to cast a spell, the player must use the mana resource. To define the cost of a spell in the system, we calculate it from the other features. Firstly, every spell type is given its own base cost. Each status has a contribution weight, and so do the status effects. However, some status effects have negative weights, to decrease the spell's cost if present in the spell features. This is to give an inherent advantage to otherwise unhelpful effects, such as spells configured to deal damage to the caster.

\subsubsection{Effects and Triggers}
To keep behavior reusable, and facilitate the mapping of status-effects functionality, we implemented a Trigger and Effect system. 
Whenever a spell is configured, the system binds the appropriate Trigger script to the spell's GameObject\footnote{Unity Technologies. "GameObject." \textit{Unity Manual}. Available: \url{https://docs.unity3d.com/6000.1/Documentation/Manual/class-GameObject.html}}. For example, it can use a \textit{HitTrigger} to cause an effect upon collision, and this \textit{HitTrigger} can be used in different types for reusability. 
Additionally, every trigger also receives a list of Effects. For example, a \textit{DamageEffect} can be used to deal damage to the target of a Trigger. Mixing and matching different triggers and effects is how our system generates the behavior for every spell. We keep the system modular through this strategy, where each piece of spell behavior is defined separately from the spell type and can be reused in different spell types, and changing the behavior of a spell is as simple as toggling a Trigger/Effect on or off.

\subsection{Challenges faced when creating the Dataset}
\begin{figure}[h]
    \centering
    \includegraphics[width=\textwidth]{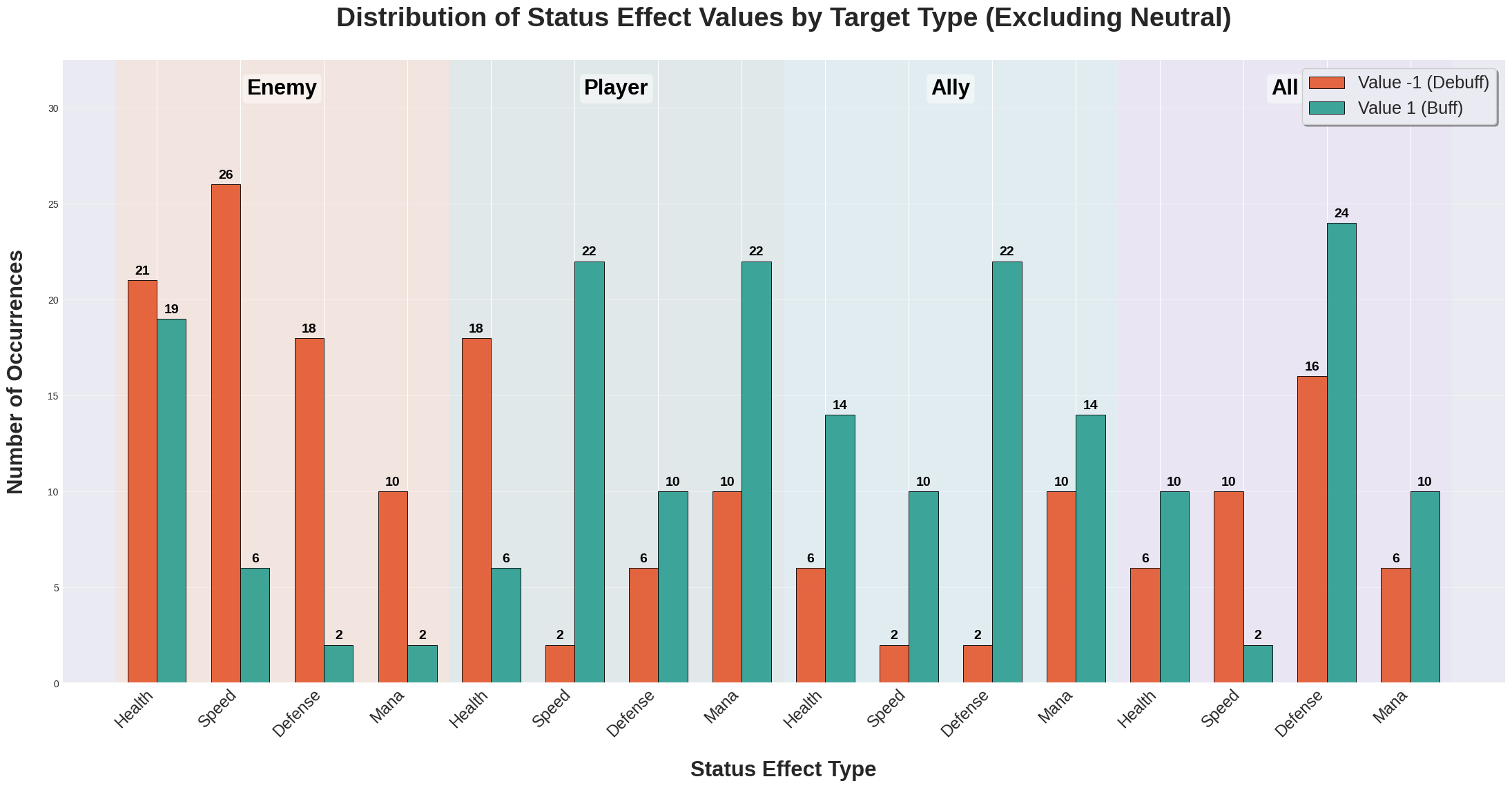}
    \caption{Distribution of the spell database, grouped by type}
    \label{fig:my_image}
\end{figure}

Our system defines features for how a spell should be represented, and as such there was no dataset available that could easily be repurposed for our goals of teaching the model. We also did not have a user base that could inspire our direction. The requirement of generating synthetic data for all features proved a challenge, especially as new features were added to the system while the model training was being built. To solve some of these issues, we utilized an automation linked to GPT-3, using a few-shot technique \cite{10.1145/3582688} by being fed initial examples that were manually created to guide new creations. We then saved these creations and used this method several times to keep a good average of examples for every possible status. We kept making sure the examples were well spread by monitoring them through graphs, like the figure \ref{fig:my_image}. However, many examples were cut due to repetitive and unhelpful generation.

\section{Results and Conclusion}

SpellForger is an ambitious project that hopes to create an intersection between Language Models and gameplay. A demonstration of the model used in the game can be found on Youtube\footnote{https://youtu.be/H0A8j9bjuiQ}. Giving players the ability to describe their own spells through natural language has great potential in allowing creativity in gameplay, replayability, and strategic personalization. A successful attempt at this system can open new paths
to AI usage in games as a tool for co-creation between player and developer, shaping truly unique and dynamic experiences in the game.

As our current methodology requires creating and maintaining a dataset, which may be a difficult task for untrained developers, future works could attempt to simplify this process further. For instance, through the construction of a tool that implements an automated pipeline powered by a generative model, where developers could merely describe the expected parameters the AI should learn and the dataset would be created or enhanced, without the need to create many examples or update them manually, thus making AI usage more easily available for this purpose.

\bibliographystyle{sbc}
\bibliography{referencias}

\end{document}

%% file: sbgames_info.tex
\trilha{MAGICA} 
\local{Salvador/BA} 
\ano{2025} 
\edicao{XXIV} 